\documentclass[10pt,twocolumn,letterpaper]{article}

\usepackage[pagenumbers]{cvpr} 
\usepackage{graphicx}
\usepackage{amsmath}
\usepackage{amssymb}
\usepackage{booktabs}
\usepackage{tabu}
\usepackage{multirow}
\usepackage{float}
\usepackage{tikz}
\PassOptionsToPackage{table, dvipsnames}{xcolor}
\usepackage[table]{xcolor}
\usepackage[dvipsnames]{xcolor}
\usepackage{colortbl}
\definecolor{pink1}{rgb}{1.0, 0.51, 0.58}
\usepackage[pagebackref,breaklinks,colorlinks]{hyperref}

\usepackage[capitalize]{cleveref}
\crefname{section}{Sec.}{Secs.}
\Crefname{section}{Section}{Sections}
\Crefname{table}{Table}{Tables}
\crefname{table}{Tab.}{Tabs.}
\definecolor{lightblue}{rgb}{0.88, 0.96, 1.0}
\definecolor{lightpurple}{rgb}{0.75, 0.5, 0.75}

\author{
\textbf{Yuqing Wen}$^{1}$\footnotemark[1]\quad 
\textbf{Hebei Li}$^{1}$\footnotemark[1]\quad 
\textbf{Kefan Gu}$^{2}$\footnotemark[1]\quad \textbf{Yucheng Zhao}$^{3}\footnotemark[2]$ \\\textbf{Tiancai Wang}$^{3}$\quad 
\textbf{Xiaoyan Sun}$^{1}$\footnotemark[3]\\
$^{1}$University of Science and Technology of China, \\  $^{2}$Nanjing University,  $^{3}$Dexmal \\
\\
\textit{Project Page}:~\href{https://wenyuqing.github.io/llada-vla/}{https://wenyuqing.github.io/llada-vla/} 
}

\begin{document}

\title{\hspace{1.7cm} LLaDA-VLA: Vision Language Diffusion Action Models}

\makeatother
\maketitle
\renewcommand{\thefootnote}{\fnsymbol{footnote}}
\footnotetext[1]{This work was done during the internship at Dexmal.}
\footnotetext[2]{Project lead.}
\footnotetext[3]{Corresponding author.}
\renewcommand{\thefootnote}{\arabic{footnote}}


\begin{abstract}
The rapid progress of auto-regressive vision-language models (VLMs) has inspired growing interest in vision-language-action models (VLA) for robotic manipulation. Recently, masked diffusion models, a paradigm distinct from autoregressive models, have begun to demonstrate competitive performance in text generation and multimodal applications, leading to the development of a series of diffusion-based VLMs (d-VLMs). However, leveraging such models for robot policy learning remains largely unexplored. In this work, we present LLaDA-VLA, the first Vision-Language-Diffusion-Action model built upon pretrained d-VLMs for robotic manipulation. To effectively adapt d-VLMs to robotic domain, we introduce two key designs: (1) a localized special-token classification strategy that replaces full-vocabulary classification with special action token classification, reducing adaptation difficulty; (2) a hierarchical action-structured decoding strategy that decodes action sequences hierarchically considering the dependencies within and across actions. Extensive experiments demonstrate that LLaDA-VLA significantly outperforms state-of-the-art VLAs on both simulation and real-world robots.
\end{abstract}

\section{Introduction}
Auto-regressive models (ARMs) have long dominated the development of vision-language models (VLMs)~\cite{llava,llavaimproved,llavaonevision,prismatic,flamingo,openflamingo}, demonstrating strong performance in multimodal understanding and text generation. Their success has naturally inspired vision-language-action models (VLAs)~\cite{openvla,openx,cogact,pai0,rt2,rth,chatvla}, where pretrained VLMs are fine-tuned on robot-specific datasets to generate actions. While effective, ARM-based VLMs rely on sequential token generation, which constrains efficiency and limits flexibility due to their inherently unidirectional generation, especially in complex multimodal robotic tasks.

Recently, masked diffusion models (MDMs)~\cite{simplemdm,discretemdm,simplifiedmdm} have emerged as a competitive alternative to auto-regressive approaches. Instead of generating tokens sequentially, MDMs produce them in parallel and iteratively refine predictions through a discrete diffusion process. A seminal work in language generation, LLaDA~\cite{llada}, demonstrates the remarkable effectiveness and scalability of this paradigm with large-scale pretraining, while subsequent studies, such as LLaDA-V~\cite{lladav} and MMaDA~\cite{mmada}, further extend this paradigm to the vision-language domain, forming diffusion-based VLMs (d-VLMs) that achieve performance comparable to leading ARM-based models. Despite this progress, their potential for robotic action generation remains unexplored, motivating our development of LLaDA-VLA, the first Vision-Language-Diffusion-Action model built upon pretrained d-VLMs.

\begin{figure}[t]
    \centering
\includegraphics[width=0.95\linewidth]{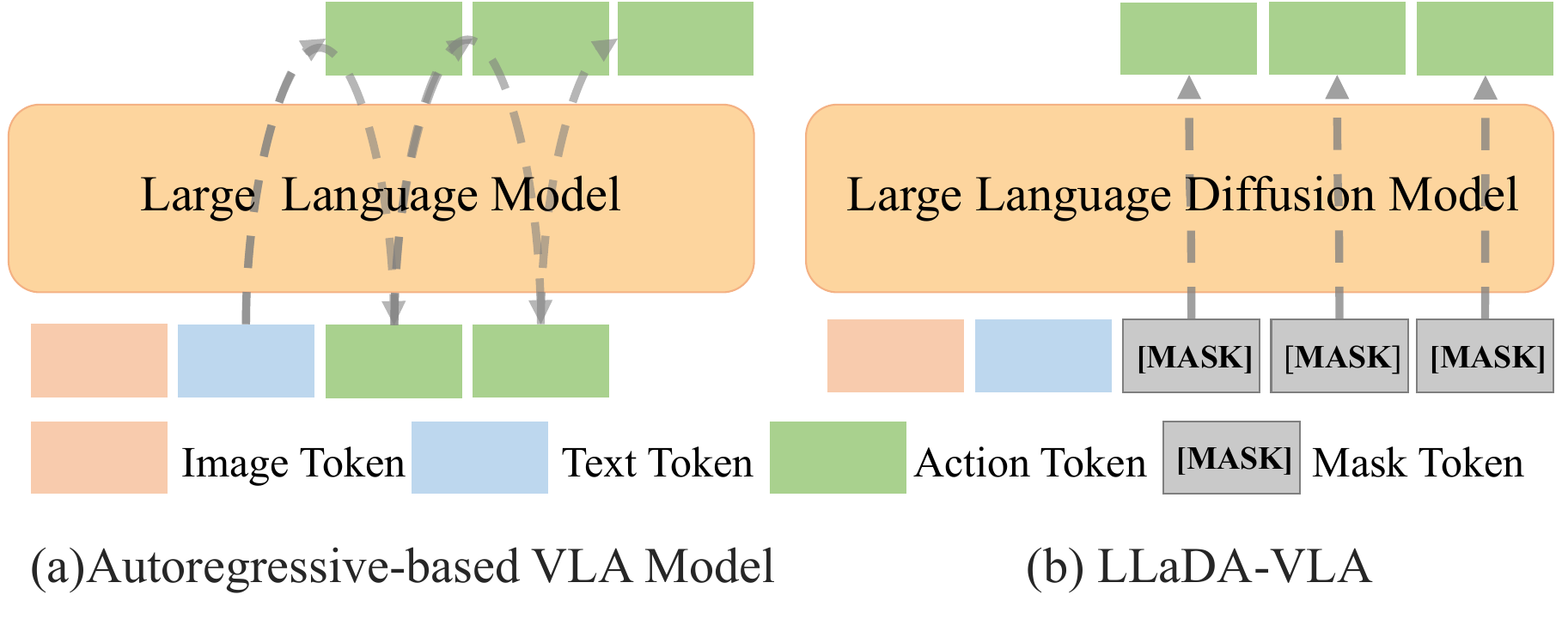} 
\captionsetup{skip=5pt}
    \caption{Comparison between Autoregressive-based VLA Model and LLaDA-VLA.}
    \vspace{-0.2cm}
    \label{fig:compare}
\end{figure}
Adapting d-VLMs to robotic tasks poses unique challenges in two aspects. First, a substantial domain gap exists: d-VLMs are trained on large-scale general datasets rich in high-level semantics, whereas VLAs must interpret low-level visual cues to generate precise actions. Second, the masked diffusion paradigm is not naturally suited for generating structured action sequences, as its decoding strategy fails to account for the structural characteristics of action sequences during decoding, making it difficult to model the strong hierarchical dependencies of robotic actions and to generate reasonable action trajectories.

To address these challenges, we propose two key strategies. First, a \textbf{localized special-token classification strategy} that mitigates the domain gap by restricting the classification space to special action tokens, significantly easing the difficulty of adaptation to robotic environments. Second, a \textbf{hierarchical action-structured decoding strategy} explicitly considers the dependencies within and across actions. Concretely, we first estimate action-level confidence scores and rank actions to determine their decoding order. Within each action, we further rank individual tokens by token-level confidence to guide decoding within the action. Together, these designs enable the masked diffusion paradigm to generate coherent and precise action sequences in robotic tasks.

LLaDA-VLA demonstrates state-of-the-art performance on both simulation and real-robot benchmarks. Compared to ARM-based baselines such as OpenVLA~\cite{openvla}, it achieves a 0.74 average length improvement on CALVIN and 51.3\% average success rate gain on SimplerEnv benchmark. LLaDA-VLA also surpasses methods such as $\pi0$~\cite{pai0} and CogACT~\cite{cogact}, achieving improvements of 23\% and 28\% in average success rate on real robots. In summary, our main contributions are:
\begin{itemize}
\item We propose the first Vision-Language-Diffusion-Action model (LLaDA-VLA) built on pretrained d-VLMs, establishing a new paradigm for robotic policy learning.
\item We design two techniques to make the masked diffusion model well-suited for action generation: a localized special-token classification strategy for easier domain adaptation, and a hierarchical action-structured decoding strategy for seamless integration into action generation.
\item Extensive experiments on simulation benchmarks SimplerEnv and CALVIN and the WidowX real robot demonstrate LLaDA-VLA's superior performance, highlighting the potential of d-VLMs for robotic manipulation.
\end{itemize}
\section{Related Work}
\subsection{Large Language Diffusion Models}
Diffusion models \cite{ddpm,sd,sd3,ddim,dit} have achieved remarkable progress in the vision domain in recent years, sparking growing interest in extending their capabilities to text generation. However, due to the inherently discrete nature of textual tokens, directly transferring diffusion models that are originally designed to operate in continuous pixel representation spaces remains challenging. To address this issue, one line of work \cite{analog,tess,dinoiser,diffuseq} proposes learning continuous representations for text, while another focuses on developing discrete diffusion models \cite{yourdiscrete,scalingdiscrete,longllada,dream7b,llada1.5}. Among them, masked diffusion models, a specific variant of discrete diffusion, have demonstrated impressive potential. Representative works such as Dream7B \cite{dream7b} and LLaDA \cite{llada} leverage large-scale language pretraining and exhibit text generation performance and scaling properties comparable to those of auto-regressive large language models.

Beyond the text domain, large language diffusion models have also demonstrated promising performance in various multimodal settings~\cite{lladav,lavida,mmada,diffa}. In the vision-language domain, LaViDA~\cite{lavida} employs a discrete diffusion transformer with multi-view image encoding and masked-denoising training. LLaDA-V~\cite{lladav} integrates masked diffusion with visual instruction tuning, enabling parallel decoding and controllable infilling. MMaDA~\cite{mmada} further proposes a unified diffusion transformer that aligns reasoning across text and vision through chain-of-thought supervision and reinforcement learning. Beyond the vision-language domain, DIFFA~\cite{diffa} extends diffusion-based LLMs to audio understanding tasks by aligning spoken content with textual representations, while LLaDA-MedV~\cite{lladamedv} adapts masked diffusion for medical applications, integrating imaging and textual data for diagnostic reasoning. Despite these advancements, applying large language diffusion models to robotic manipulation remains largely unexplored, leaving significant avenues for future research.

\subsection{Vision-Language-Action Models}
Building a generalist robotic policy\cite{octo,peract,robofg,rvt,rvt2,imagebc,C2FARM,rpt,vima,rt1,palme} has long been a challenging and highly desired goal. In recent years, the rapid advances in auto-regressive Vision-Language Models (VLMs)\cite{llava,llavaimproved,llavaonevision,prismatic,flamingo} have inspired researchers to leverage their strong multimodal understanding and generalization capabilities to develop robotic policies, commonly referred to as Vision-Language-Action (VLA) models~\cite{cogact,pai0,rt2,chatvla,rth,openvla,cot,whatmatters,fastecot}. Among them, RT-2~\cite{rt2} is a pioneering work that jointly fine-tunes pretrained VLMs on a combination of web-scale VQA data and robot demonstrations, achieving strong multi-task performance across diverse manipulation tasks. Building upon this idea, OpenVLA~\cite{openvla} is introduced as the first open-source VLA model, further promoting research progress in this area. Subsequently, LLARVA~\cite{llarva} improves action prediction by incorporating trajectory annotations; CogACT~\cite{cogact} introduces a large diffusion action head to predict continuous control commands; and $\pi0$~\cite{pai0} adopts a flow-matching strategy combined with carefully curated large-scale multi-task datasets, achieving remarkable performance.

Despite these advances, existing VLA models are almost exclusively built on auto-regressive VLMs, leaving the potential of diffusion-based VLMs largely unexplored. In this work, we investigate how to build a Vision-Language-Diffusion-Action model based on pretrained d-VLMs. A concurrent work~\cite{discretevla} shares partial inspiration with us , but their approach still relies on auto-regressive VLMs.

\begin{figure*}[t]
    \centering
    \includegraphics[width=1.0\textwidth]{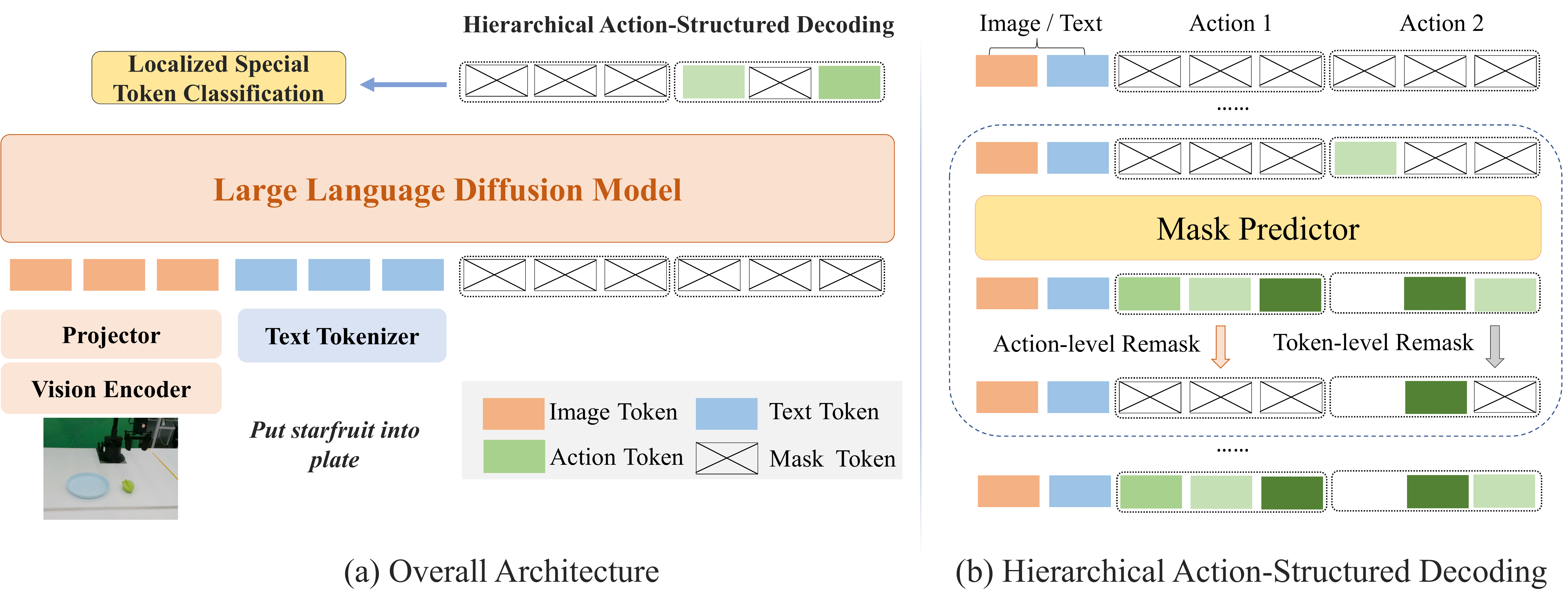} 
\captionsetup{skip=5pt}
\caption{\textbf{Overview of LLaDA-VLA}. (a) Overall architecture. Visual features extracted by the vision encoder are projected into the text space and concatenated with text tokens. Together with masked tokens, they are fed into a large language diffusion model to generate action sequences via Localized Special-Token Classification and further refined with Hierarchical Action-Structured Decoding.
(b) Hierarchical Action-Structured Decoding strategy. Starting from a fully masked action sequence (except vision and text prompts), the model iteratively predicts masked tokens, performing action-level and token-level remasking based on confidence until the full sequence is decoded.}

    \label{fig:method}
\end{figure*}
\section{Method}

We present the methodology of LLaDA-VLA, starting with a brief overview of the masked diffusion model in Section~\ref{sec:pre}, followed by the architecture and key designs of LLaDA-VLA in Section~\ref{sec:llada-vla}.

\subsection{Preliminary: Mask Diffusion Models} \label{sec:pre}
Masked diffusion models (MDMs)~\cite{yourdiscrete,scalingdiscrete,llada,dream7b} define a generative paradigm based on a forward–reverse diffusion process over discrete tokens, fundamentally different from auto-regressive models. Within this paradigm, large language diffusion models such as LLaDA have demonstrated competitive performance comparable to auto-regressive counterparts. In the forward process, given an input sequence $\boldsymbol{x}_{0}=\left[\boldsymbol{x}_0^{i}\right]_{i=1}^{N}, \boldsymbol{x}_0^{i}  \in \{0,1,\dots,\mathcal{V}-1\}^N$ of length $N$ with vocabulary size $\mathcal{V}$, each token in $\mathbf{x}_0$ is independently replaced by a special mask token $\mathrm{[M]}$ with probability $t$, formulated as:
\begin{equation}q_{t \mid 0}\left(\boldsymbol{x}_{t} \mid \boldsymbol{x}_{0}\right)=\prod_{i=0}^{N-1} q_{t \mid 0}\left(\boldsymbol{x}_{t}^{i} \mid \boldsymbol{x}_{0}^{i}\right),\end{equation}

\begin{equation} q_{t \mid 0}\left(\boldsymbol{x}_{t}^{i} \mid \boldsymbol{x}_{0}^{i}\right)=\left\{\begin{array}{ll}
1-t, & \boldsymbol{x}_{t}^{i}=\boldsymbol{x}_{0}^{i} \\
t, & \boldsymbol{x}_{t}^{i}=[\mathrm{M}] .
\end{array}\right.\end{equation}

In a reverse process, MDMs gradually transform masked tokens into meaningful content, starting from a fully masked sequence. Given $0 < s < t < 1$, each sampling step in the reverse process can be formulated as:
\begin{equation}q_{s \mid t}\left(\boldsymbol{x}_{s} \mid \boldsymbol{x}_{t}\right)=\prod_{i=0}^{N-1} q_{s \mid t}\left(\boldsymbol{x}_{s}^{i} \mid \boldsymbol{x}_{t}\right),\end{equation}

\begin{equation}q_{s \mid t}\left(\boldsymbol{x}_{s}^{i} \mid \boldsymbol{x}_{t}\right)=\left\{\begin{array}{ll}
1, & \boldsymbol{x}_{t}^{i} \neq[\mathrm{M}], \boldsymbol{x}_{s}^{i}=\boldsymbol{x}_{t}^{i} \\
\frac{s}{t}, & \boldsymbol{x}_{t}^{i}=[\mathrm{M}], \boldsymbol{x}_{s}^{i}=[\mathrm{M}], \\
\frac{t-s}{t} p_{\boldsymbol{\theta}}\left(\boldsymbol{x}_{0}^{i} \mid \boldsymbol{x}_{t}\right), & \boldsymbol{x}_{t}^{i}=[\mathrm{M}], \boldsymbol{x}_{s}^{i} \neq[\mathrm{M}], \\
0, & \text { otherwise },
\end{array}\right.\end{equation}
where $p_{\boldsymbol{\theta}}$ is modeled by a Transformer as mask predictor. This means that for each reverse sampling step, a fraction of $(1-s/t)$ tokens are predicts through mask predictor $p_{\boldsymbol{\theta}}(\boldsymbol{x}_{0}^{i} \mid \boldsymbol{x}_{t})$, while a fraction $s/t$ of tokens are remasked and deferred for re-predict in subsequent sampling steps. A typical strategy, as adopted in LLaDA, is to choose the $s/t$ tokens for remask with the lowest confidence scores (i.e., the smallest logits). 

The mask predictor $p_{\boldsymbol{\theta}}$ is trained to predict mask tokens with cross-entropy loss only computed on masked tokens: 
\begin{equation}\mathcal{L}(\theta) \triangleq-\mathbb{E}_{t, x_{0}, x_{t}}\left[\frac{1}{t} \sum_{i=1}^{L} \mathbf{1}\left[x_{t}^{i}=\mathbf{M}\right] \log p_{\theta}\left(x_{0}^{i} \mid x_{t}\right)\right]\end{equation}

\subsection{LLaDA-VLA}\label{sec:llada-vla}
In this section, we provide a detailed description of LLaDA-VLA. We first introduce the overall model architecture, including the structure of our VLA framework and the processing of its inputs and outputs. We then present two key designs proposed in this work: localized special-token classification and hierarchical action-structured decoding strategy.

\subsubsection{Model Architecture}

\textbf{Vision-language Modules: }As shown in Figure.~\ref{fig:method}~(a), LLaDA-VLA consists of three main components: a language backbone, a vision encoder, and a projector. Following LLaDA-V~\cite{lladav}, we use LLaDA~\cite{llada} as the language backbone, and adopt SigLIP-2~\cite{siglip2} as the vision encoder and MLP as projector, respectively. The model takes two inputs: a language instruction that specifies the robot’s task and a front-view RGB image. The vision encoder processes the image to extract visual features, which are then projected by the MLP into the shared space with the text tokens. The visual and text tokens are then concatenated and fed into the large language diffusion model to generate the robot action sequence in a manner that is detailed in Section~\ref{sec:LSC} and Section~\ref{sec:HAD}.

\noindent\textbf{Action Tokenization and Chunking: }To enable the language model to generate robot actions, we discretize continuous action values into bins with bin size of $\mathcal{V}_a$. We augment the original vocabulary with $\mathcal{V}_a$ additional special tokens  $\mathcal{S} = \{s_0, s_1, \dots, s_{\mathcal{V}_a-1}\}$, where $\mathcal{V}_a \ll \mathcal{V}$, to represent these discrete action tokens, resulting in a vocabulary size of $\mathcal{V}_{total} = \mathcal{V} + \mathcal{V}_a$. Therefore, a per-timestep action is represented by $D=7$ special action tokens: three for positional displacements, three for rotational changes, and one for the gripper open/close state. To generate multi-step trajectories, the model predicts an action chunk spanning $K$ consecutive timesteps, yielding a robot action sequence of $K \times D$ special action tokens. These tokens can be de-tokenized to recover the original continuous values, allowing the model to produce executable robot trajectories.

\subsubsection{Localized Special-token Classification}\label{sec:LSC}
The original training objective of pretrained d-VLMs is to perform full-vocabulary classification for each token. To generate actions, however, the model is required to predict only the special action tokens. Therefore, retaining the full-vocabulary classification objective would complicate the learning process and hinder the adaptation of pretrained d-VLMs to robotic action generation. To address this, we introduce a localized special-token classification mechanism. 
Instead of performing classification over the entire vocabulary with size $\mathcal{V}_{total}$, we focus the classification only on the special action tokens $\mathcal{S}$. During training, each original token label $y_i \in \mathcal{V}$ is mapped to a local class $l_i \in \{0, \dots, \mathcal{V}_a-1\}$ as:
\begin{equation}
l_i =
\begin{cases}
\text{map}(y_i), & \text{if } y_i \in \mathcal{S},\\
-100, & \text{otherwise (ignored in loss)},
\end{cases}
\end{equation}
where $\text{map}(\cdot)$ denotes the mapping from the original token index in the full vocabulary to the local class index. We then choose only the logits on the special action tokens
$z_i = \text{logits}[i, \mathcal{S}] \in \mathbb{R}^{\mathcal{V}_a}$ and compute the token-level cross-entropy loss only on masked positions with the training objective:
\begin{equation}
L_\text{token} = \frac{1}{|M|} \sum_{i \in M} \text{CE}(z_i, l_i),
\end{equation}
where $\mathrm{CE}(\cdot)$ denotes the standard cross-entropy loss and M is the set of valid masked positions. During inference, we predict over the target token subset $\mathcal{S}$ only, and map the predicted local class indexes back to the original token indexes to get the action tokens. This localized classification objective concentrates learning on action-relevant tokens, thereby improving the accuracy for action generation and making the training easier.

\subsubsection{Hierarchical Action-Structured Decoding}\label{sec:HAD}
In the original LLaDA decoding process, as illustrated in Figure.~\ref{fig:method}~(b), starting from a fully masked sequence, the model first predicts the masked tokens and retains those with high confidence, while remasks the remaining tokens. In the subsequent diffusion step, the remasked tokens are re-predicted, and again a fraction of tokens with high confidence are retained and others remasked. By iteratively repeating this predict–remask–predict process, the model gradually refines its generated output until obtaining the desired result. 

However, this conventional decoding strategy cannot be directly applied to generating action chunks. This is primarily because it treats all output tokens equally, ignoring the structured dependencies between them. In contrast, an action chunk exhibits both intra-action and inter-action correlations. 
To explicitly capture such correlations, we introduce a hierarchical action-structured decoding strategy that accounts these dependencies during generation. We first compute \emph{action-level confidence scores} for each action by summing the confidence of tokens within each action:
\begin{equation}
C_a^{(i)} = \sum_{j=1}^{D} c_{i,j},
\end{equation}
where $c_{i,j}$ denotes the confidence of the $j$-th token within the $i$-th action, and $D$ is the number of tokens per action. At each hierarchical decoding step, we first rank all actions within the predicted action chunk according to their action-level confidence scores $C_a^{(i)}$. The action with the highest confidence $C_a$ is selected for partial preservation and the rest of the actions are remasked (action-level remask). Within this selected action, its tokens are further ranked by a token-level confidences. Only a subset of high-confidence tokens is retained with the rest remasked (token-level remask). The remasked tokens are then regenerated in subsequent diffusion steps. This hierarchical decoding procedure ensures that the trajectory is generated in an action-wise manner, preserving the structural integrity while allowing further refinement within each individual action. In this way, the model can effectively generate more coherent and reasonable action trajectories.

\section{Experiment}
\subsection{Experiment Setup}
\label{sec:expsetup}
\subsubsection{Dataset}
We evaluate LLaDA-VLA in the simulation environments SimplerEnv~\cite{simpler} and CALVIN~\cite{calvin}, as well as on a real-world WidowX robot.

\begin{table}[!t]
\centering
\small
\caption{Performance comparison on the WidowX robot in the SimplerEnv \textit{Visual Matching} setting. We compare success rates (\%) on 4 tasks. LLaDA-VLA achieves the best average success rate.}
\label{tab:simpler}
\resizebox{0.48\textwidth}{!}{
\begin{tabular}{l|c|c|c|c|c}
\toprule
Method &
  \begin{tabular}[c]{@{}c@{}}Put Spoon\\ on Towel\end{tabular} &
  \begin{tabular}[c]{@{}c@{}}Put Carrot\\ on Plate\end{tabular} &
  \begin{tabular}[c]{@{}c@{}}Stack Green\\ on Yellow\end{tabular} &
  \begin{tabular}[c]{@{}c@{}}Put Eggplant\\ in Basket\end{tabular} &
  Average \\
\midrule
RT-1-X~\cite{openx}         & 0.0   & 4.2   & 0.0   & 0.0    & 1.1   \\ 
Octo-Base~\cite{octo}               & 15.8  & 12.5  & 0.0   & 41.7   & 17.5  \\ 
Octo-Small~\cite{octo}              & 41.7  & 8.2   & 0.0   & 56.7   & 26.7  \\ 
OpenVLA~\cite{openvla}               & 4.2   & 0.0   & 0.0   & 12.5   & 4.2   \\
Cog-ACT~\cite{cogact}                                      & 71.7  & 50.8  & 15.0  & 67.5   & 51.3  \\
DiscreteDiffusionVLA~\cite{discretevla}                          & 37.5  & -   & 20.8  & 29.2   & 29.2  \\
 \rowcolor[gray]{.9} 
LLaDA-VLA               & 56.9  & 76.3  & 30.6  & 58.3   & 55.5\\
\bottomrule
\end{tabular}}
\label{tab:widowx_full}
\end{table}

\begin{table}[t] 
\centering
\caption{Comparison of methods on CALVIN ABC-D setting. We report the average success rate over 1000 rollouts per task, along with the average number of tasks completed consecutively to accomplish five instructions (Avg. Len.). LLaDA-VLA consistently outperforms previous methods.}
\label{tab:calvin}
\resizebox{0.48\textwidth}{!}{
\begin{tabular}{l|ccccc|c}
\toprule
Method & \multicolumn{5}{c|}{Task completed in a row} & Avg. Len. $\uparrow$ \\
\cmidrule(lr){2-6}
 & 1 & 2 & 3 & 4 & 5 & \\
\midrule
Roboflamingo~\cite{robofg} & 82.4 & 61.9 & 46.6 & 33.1 & 23.5 & 2.47 \\
Susie~\cite{susie} & 87.0 & 69.0 & 49.0 & 38.0 & 26.0 & 2.69 \\
GR-1~\cite{gr1} & 85.4 & 71.2 & 59.6 & 49.7 & 40.1 & 3.06 \\
3D Diffusor Actor~\cite{3dactor} & 92.2 & 78.7 & 63.9 & 51.2 & 41.2 & 3.27 \\
OpenVLA~\cite{openvla} & 91.3 & 77.8 & 62.0 & 52.1 & 43.5 & 3.27 \\
 \rowcolor[gray]{.9} LLaDA-VLA & 95.6 & 87.8 & 79.5 & 73.9 & 64.5 & 4.01 \\
\bottomrule
\end{tabular}
}
\end{table}

\begin{table*}[tb]
\centering
    \caption{Comparison with previous methods on real robot. We compare the success rates (\%) across four tasks. LLaDA-VLA achieves the best performance.}
    \label{tab:real_in}
\resizebox{1.0\textwidth}{!}{
\begin{tabular}{lccccc}
\hline
\textbf{Method} & \textbf{Banana on Plate} & \textbf{Strawberry in Bowl} & \textbf{Starfruit on Plate} & \textbf{Banana\&Strawberry in Bowl} & \textbf{Average Success Rate} \\
\hline
$\pi_0$~\cite{pai0}        & 50\% & 30\% & 40\% & 20\% & 35\% \\
CogACT\cite{cogact}        & 40\% & 30\% & 30\% & 20\% & 30\% \\
 \rowcolor[gray]{.9} LLaDA-VLA        & 50\% & 70\% & 70\% & 40\% & 58\% \\

\hline
\end{tabular}}
\label{tab:success_rates}
\end{table*}

\begin{table*}
\caption{Performance on unseen tasks for real robot evaluating the generalization capability. Comparing the success rate across these OOD tasks, LLaDA-VLA shows better performance than $\pi0$.}
    \centering
    \footnotesize
        \centering
    \label{tab:unseen}
\resizebox{1.0\textwidth}{!}{        \begin{tabular}{lccccc}
            \toprule
              \textbf{Method} & \textbf{Cube on Plate}&\textbf{Strawberry in Box}&\textbf{Cube in Box}&\textbf{Banana\&Strawberry in Bowl(dist)}& \textbf{Average Success Rate} \\
              \hline
             $\pi$0~\cite{pai0}  &30&20&10&0&15\\
             \rowcolor[gray]{.9} LLaDA-VLA  &50&60&50&0&40\\
            \bottomrule
        \end{tabular}
        }
\end{table*}

\begin{table}[t] 
\centering
\caption{Ablation on the localized special-token classification(LSC) and the hierarchical
action-structured decoding(HAD) on CALVIN ABC-D setting.}
\label{tab:ablation}
\resizebox{0.4\textwidth}{!}{ 
\begin{tabular}{l|ccccc|c}
\toprule
Method & \multicolumn{5}{c|}{Task completed in a row} & Avg. Len. $\uparrow$ \\
\cmidrule(lr){2-6}
 & 1 & 2 & 3 & 4 & 5 & \\
\midrule
baseline  & 86.2 & 62.4 & 46.1 & 34.1 & 24.7& 2.64\\
+ LSC & 91.6 & 80.1 & 67.2 & 57.3 & 46.4 & 3.43(\textcolor{blue}{+0.79})\\
+ HAD & 95.6 & 87.8 & 79.5& 73.9 & 64.5 & 4.01(\textcolor{blue}{+0.58}) \\
\bottomrule
\end{tabular}}
\end{table}

\begin{table}[t] 
\centering
\caption{Ablation on the action chunk size on CALVIN ABC-D setting.}
\label{tab:chunk}
\resizebox{0.4\textwidth}{!}{ 
\begin{tabular}{l|ccccc|c}
\toprule
Method & \multicolumn{5}{c|}{Task completed in a row} & Avg. Len. $\uparrow$ \\
\cmidrule(lr){2-6}
 & 1 & 2 & 3 & 4 & 5 & \\
\midrule
3  & 95.0 & 86.8 & 78.2 & 70.4 & 59.9& 3.90\\
\rowcolor[gray]{.9}
5 & 95.6 & 87.8 & 79.5 & 73.9 & 64.5 & 4.01 \\
8 & 91.1 & 79.5 & 69.0 & 61.3 & 52.4& 3.53\\
10 & 90.8 & 76.5 & 65.8 & 55.7 & 46.0 & 3.36\\

\bottomrule
\end{tabular}}
\end{table}

\begin{figure}[htbp]
    \centering
    \includegraphics[width=0.4\textwidth]{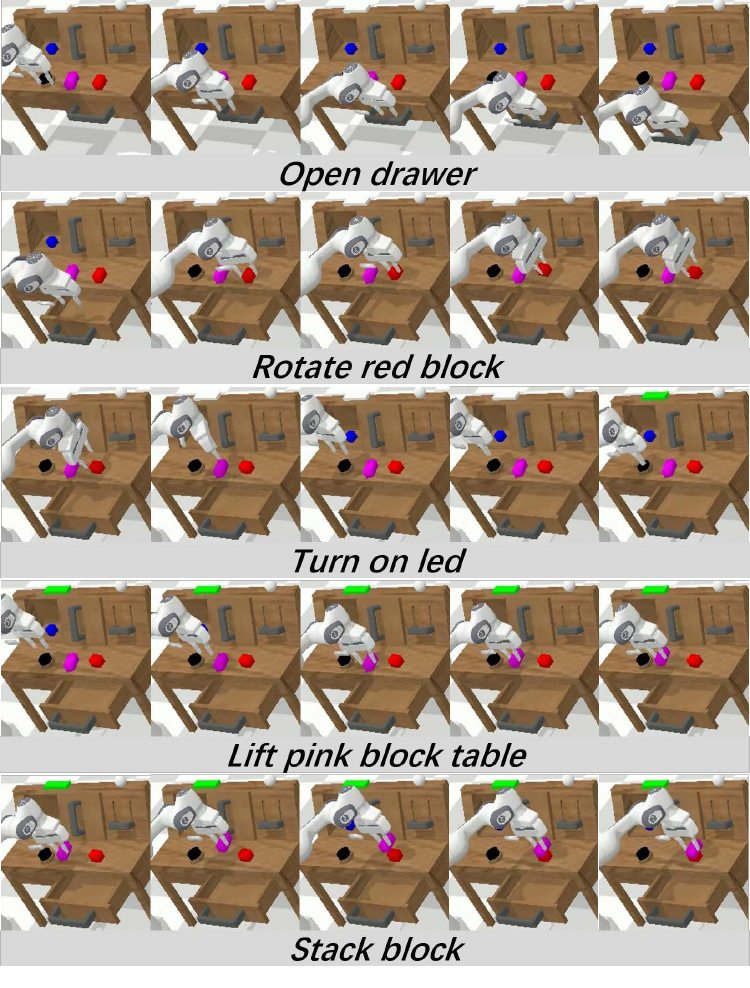}
    \caption{Qualitative results of LLaDA-VLA on CALVIN tasks.}
    \label{fig:calvin_exp}
\end{figure}

\begin{figure*}[htbp]
    \centering
    \includegraphics[width=0.8\textwidth]{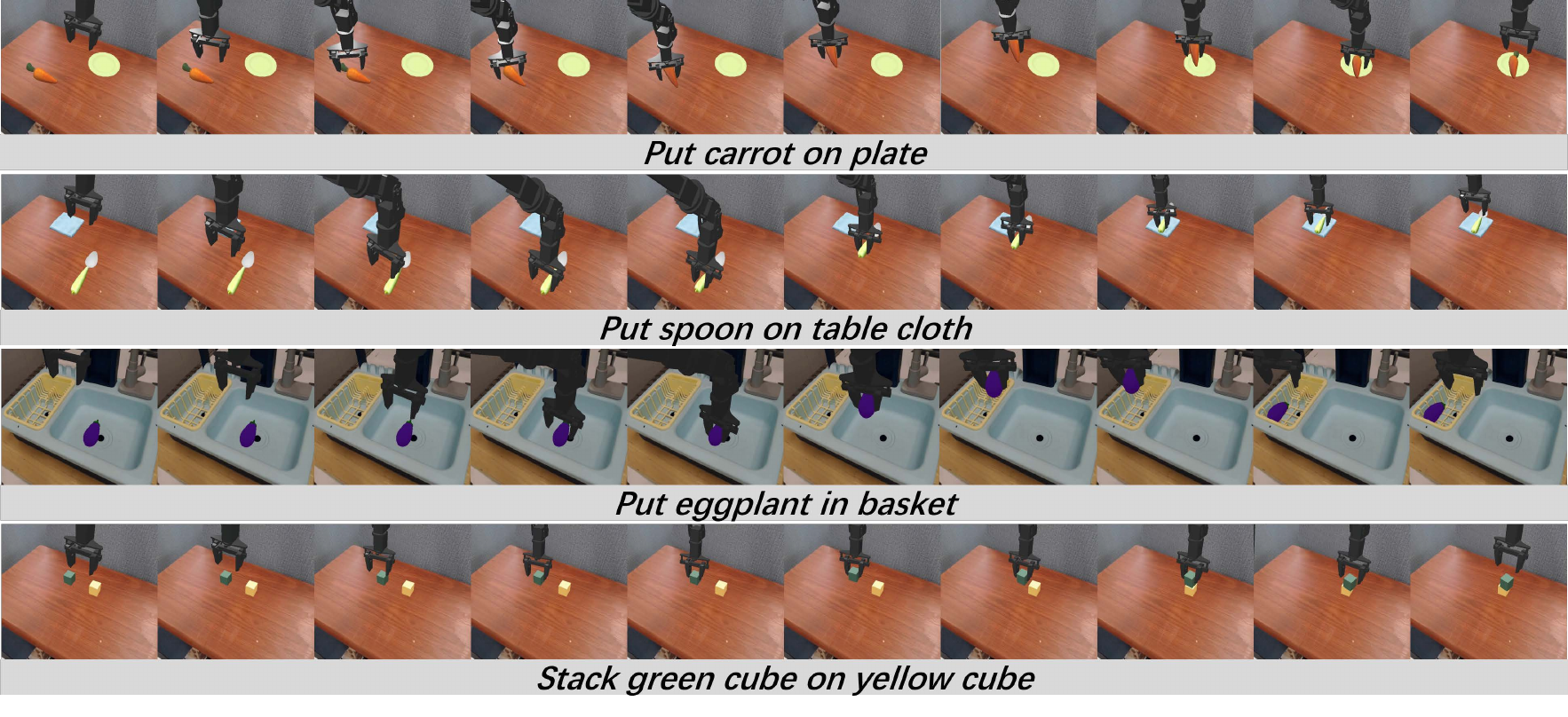}
    \caption{Qualitative results of LLaDA-VLA on SimplerEnv tasks.}
    \label{fig:simpler_exp}
\end{figure*}

\begin{figure*}[htbp]
    \centering
    \includegraphics[width=0.8\textwidth]{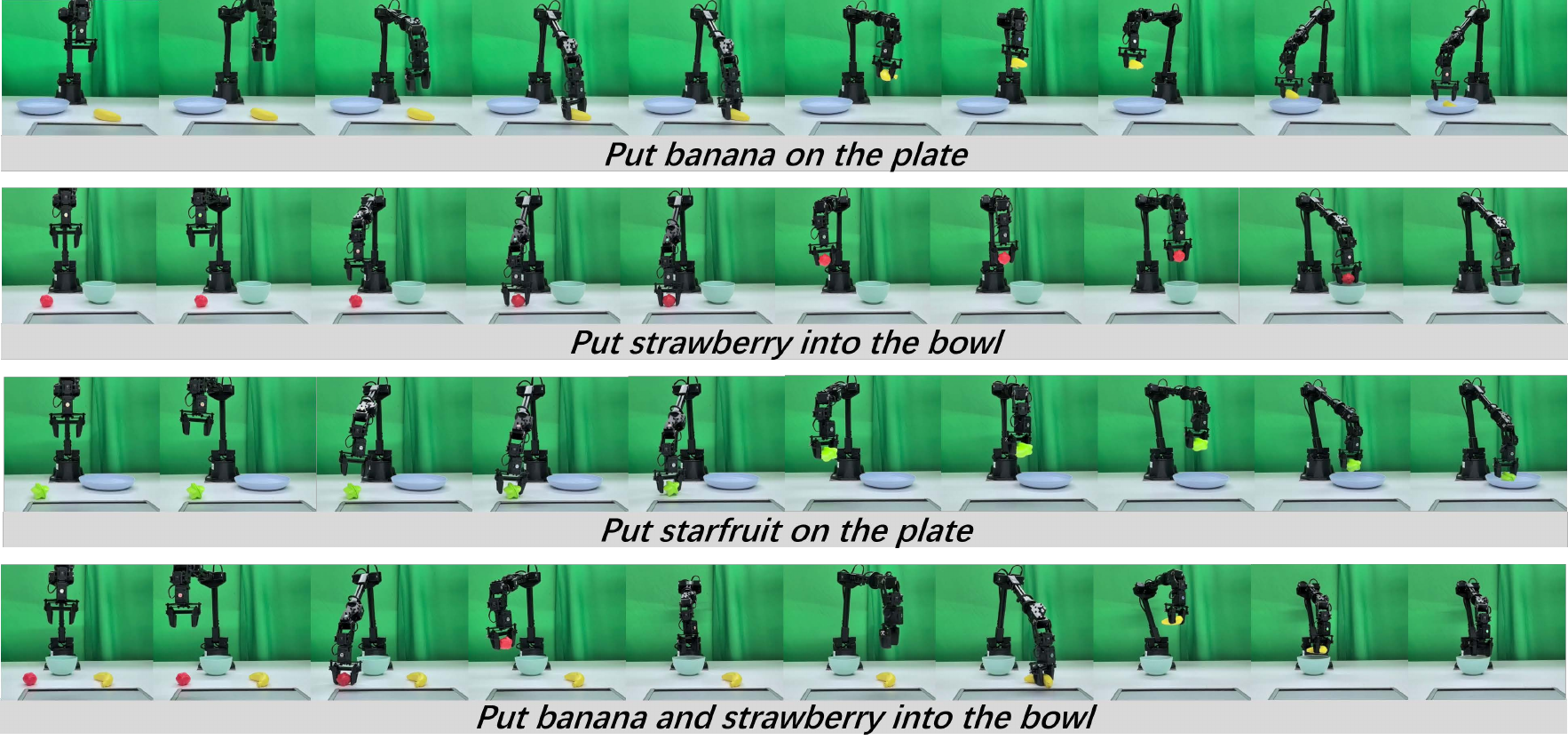}
    \caption{Qualitative results of LLaDA-VLA on real-world in-domain tasks.}
    \vspace{-0.2cm}
    \label{fig:real_id_exp}
\end{figure*}

\begin{figure*}[htbp]
    \centering
    \includegraphics[width=0.8\textwidth]{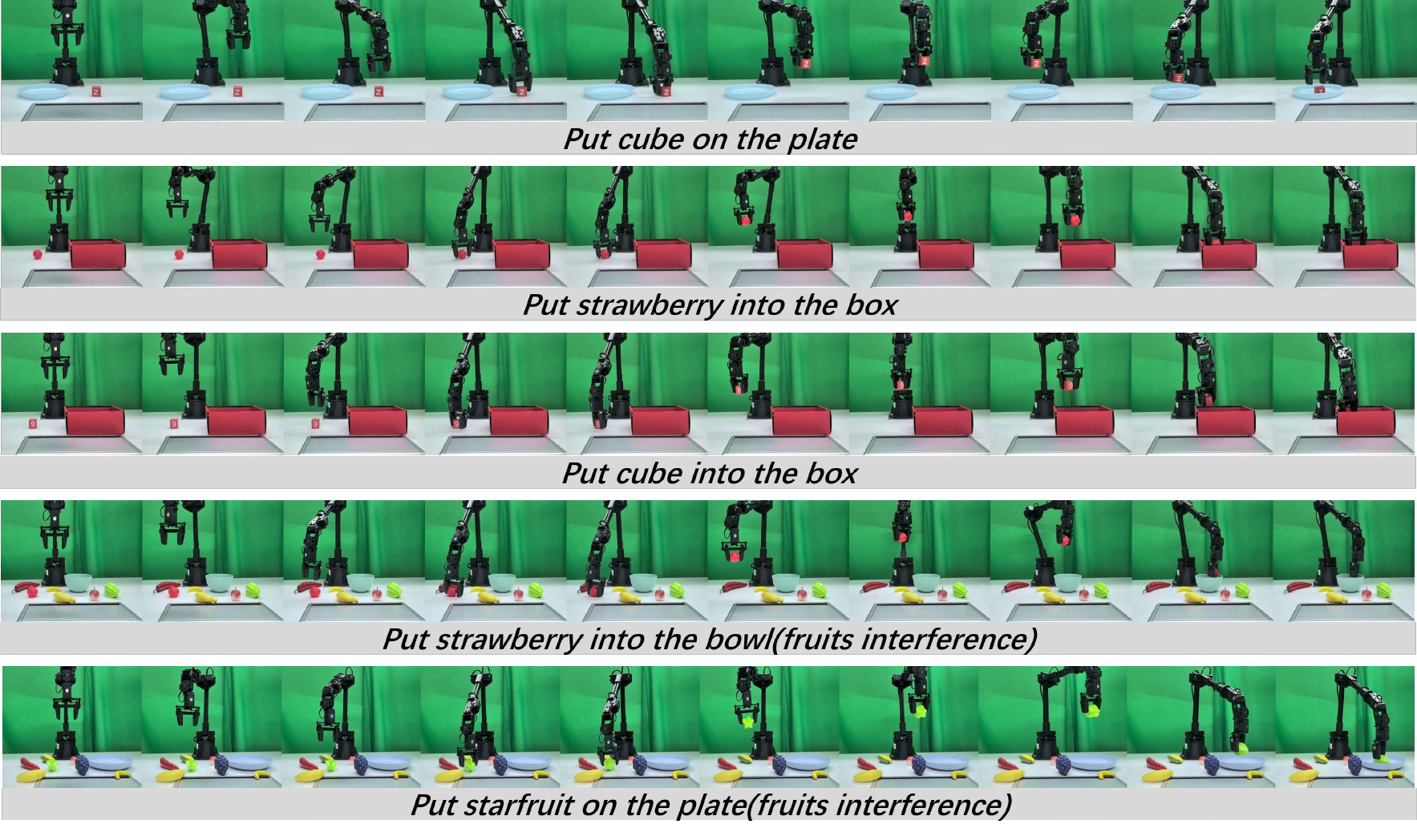}
    \caption{Qualitative results of LLaDA-VLA on real-world out-of-domain tasks.}
    \vspace{-0.2cm}
    \label{fig:real_ood_exp}
\end{figure*}
\noindent\textbf{SimplerEnv}. SimplerEnv~\cite{simpler} is designed to closely mirror real-world physical dynamics and visual appearances, providing a more faithful evaluation of robotic policies. In our experiments, we evaluate on the WidowX robot under the SimplerEnv Visual Matching setting, which minimizes the gap between simulation and reality, thereby closely approximating real-world conditions. We focus on four tasks: \textit{Put Spoon on Towel}, \textit{Put Carrot on Plate}, \textit{Stack Green on Yellow}, and \textit{Put Eggplant in Basket}. Task success rate is used as the primary evaluation metric.

\noindent\textbf{CALVIN}. 
We train and evaluate LLaDA-VLA on CALVIN, an open-source benchmark for long-horizon, language-conditioned robotic manipulation. CALVIN comprises four distinct environments (A, B, C, D) and provides diverse sensory inputs commonly used in visuomotor control, including RGB-D observations from fixed and gripper-mounted cameras, proprioceptive states, and vision-based tactile feedback. Following the ABC-D protocol, we train on environments A, B, and C, and evaluate generalization on environment D. We report both success rate and average episode length over five consecutive tasks.

\noindent\textbf{Real-World WidowX Robot}. For real-world experiments, we use a WidowX 250S robotic arm equipped with an Intel RealSense D435 camera positioned in front of the setup to capture third-person visual observations. LLaDA-VLA is evaluated on eight tasks, comprising four seen tasks and four generalization tasks. The seen tasks include three short-horizon manipulations—\textit{Banana on Plate}, \textit{Strawberry in Bowl}, and \textit{Starfruit on Plate}—and one long-horizon task, \textit{Banana and Strawberry in Bowl}. The generalization tasks involve \textit{Cube on Plate} (introducing an unseen object), \textit{Strawberry in Box} (featuring an unseen container), \textit{Cube in Box} (involving both an unseen object and container), and a long-horizon OOD task, \textit{Banana and Starfruit in Bowl} (performed in the presence of distractor objects). Each task is executed over 10 independent trials.

\subsubsection{Training and Evaluation Details}
We use LLaDA-V\cite{lladav} as our pretrained weights, an open-source d-VLM. All our experiments are fine-tuned for 3 epochs, with a learning rate of 2e-5 and a batch size of 128. We introduce 32 additional special tokens into the vocabulary for classification. Since we adopt a fixed-length output setting, we remove the EOS token. In our main experiments, the action chunk size is set to 5, and the model predicts delta actions. During inference, we use 10 iterative diffusion steps, with 2 iterations per action. We adopt the dllm-cache~\cite{dllmcache} method to accelerate the decoding.

\subsection{Quantitative Results} 
\subsubsection{Performance Comparison}
\noindent\textbf{SimplerEnv.} We train and evaluate LLaDA-VLA on SimplerEnv, as shown in Table~\ref{tab:simpler}. LLaDA-VLA achieves an average performance of 55.1, which is significantly higher than OpenVLA, a typical
autoregressive VLA model, with 50.9\% improvements and also surpasses more advanced methods such as CogAct with 4.2 points gain. Such superior performance validates the effectiveness of the LLaDA-VLA framework.

\noindent\textbf{CALVIN.} 
We further compare LLaDA-VLA with several representative works on another widely adopted simulation benchmark, CALVIN, as summarized in Table \ref{tab:calvin}. In particular, LLaDA-VLA outperforms OpenVLA with a notable performance improvement of 0.74. It also demonstrates superior performance compared to other prominent approaches, such as GR1\cite{gr1} and RoboFlamingo~\cite{robofg}. These results provide compelling evidence of LLaDA-VLA’s effectiveness, highlighting the promising value of incorporating the masked diffusion paradigm to the VLA framework.

\noindent\textbf{Real Robot.} Beyond simulation, we further evaluate LLaDA-VLA on real-robot experiments, comparing it with state-of-the-art approaches $\pi0$ and CogACT. As shown in Table \ref{tab:real_in}, LLaDA-VLA achieves an average success rate of 58\% across four real-world manipulation tasks, including up to 70\% success rates on the \textit{Strawberry in Bowl} and \textit{Starfruit on Plate} tasks. These results consistently outperform both CogACT and $\pi0$, demonstrating the strong efficacy of LLaDA-VLA on real world setting.
\subsubsection{Generalization Capability}
We evaluate the generalization capability of LLaDA-VLA on four out-of-distribution (OOD) real-robot tasks, which involve unseen objects, containers, and distractors. As shown in Table \ref{tab:unseen}, LLaDA-VLA demonstrates strong generalization performance, achieving, for example, a 60\% success rate on \textit{Strawberry in Box} task, which is much higher than the success rate of $\pi0$. LLaDA-VLA attains a gain of 25\% average success over $\pi0$, further highlighting its outstanding generalization ability.

\subsubsection{Ablation Study}
To validate the effectiveness of our proposed designs, we conduct ablation studies on the two core components of LLaDA-VLA: the localized special-token classification mechanism and the hierarchical action-structured decoding strategy. Additionally, we investigate how the length of the generated action sequences affects performance under the diffusion paradigm. All experiments in this section are carried out on the CALVIN benchmark.

\noindent\textbf{Localized Special-token Classification}. As shown in Table~\ref{tab:ablation}, we first establish a baseline model without the localized special-token classification or the hierarchical action-structured decoding. This baseline yields suboptimal results, with an average episode length of only 2.54 and a success rate of 86.2\% on the first task, highlighting the significant challenges of directly adapting vanilla d-VLMs to robotic manipulation. When we incorporate the localized special-token classification strategy into the baseline, the performance improves substantially by 0.79. This demonstrates that focusing the model on action-specific tokens effectively reduces the adaptation difficulty, transforming the full-vocabulary classification problem into a more tractable task and enabling more efficient training.

\noindent\textbf{Hierarchical Action-structured Decoding strategy}.
We investigate the effectiveness of the hierarchical action-structured decoding strategy. As shown in Table~\ref{tab:ablation}, incorporating this decoding approach brings a substantial improvement of 0.58 points over the vanilla decoding strategy. This significant gain demonstrates that explicitly modeling both intra-action and inter-action dependencies through hierarchical decoding effectively benefit the diffusion model’s outputs, enabling the generation of more coherent action trajectories and markedly improving task success rates.

\noindent\textbf{Action Chunk Size}.
The action chunk size has a significant impact on the model’s performance for generating trajectories of varying lengths. As shown in Table~\ref{tab:chunk}, moderately increasing the chunk size can improve task success rates, as generating longer segments at once encourages smoother motion. However, when the chunk size becomes too large, the mask prediction task becomes more difficult due to the increased number of tokens to be generated, which may reduce overall accuracy. Therefore, for LLaDA-VLA, it is crucial to select an appropriate action chunk size to balance trajectory smoothness and prediction accuracy, to achieve optimal performance.

\subsection{Qualitative Results} 
In this section, we present visualizations of several tasks in both simulation and real-world, demonstrating LLaDA-VLA’s ability to accomplish a wide range of robotic tasks.

\noindent\textbf{Simulation results.}
In CALVIN, LLaDA-VLA successfully completes multi-step manipulation tasks as shown in Figure~\ref{fig:calvin_exp}. This demonstrates LLaDA-VLA’s capability to complete long-horizon tasks. In SimplerEnv, as illustrated in Figure~\ref{fig:simpler_exp}, LLaDA-VLA accurately localizes target objects and reliably grasps and places them into the correct locations, showcasing its good performance. Overall, these results demonstrate that LLaDA-VLA is capable of stably executing long-horizon tasks while also achieving precise object-level manipulation. 

\noindent\textbf{Real-world results.}
On in-domain tasks, such as \textit{Banana on Plate} or \textit{Strawberry in Bowl}, the robot reliably executes actions, as shown in Figure~\ref{fig:real_id_exp}. LLaDA-VLA also exhibits strong generalization to out-of-domain tasks, as illustrated in Figure~\ref{fig:real_ood_exp}. It is capable of grasping previously unseen objects, such as cubes, and placing items into novel targets, such as a paper box. Even in the presence of multiple distractor objects, LLaDA-VLA successfully completes tasks, for example, accurately picking and placing a strawberry despite surrounding fruits. These qualitative results demonstrate the versatility and robustness of LLaDA-VLA in real-world scenarios.

\section{Conclusion}
We present LLaDA-VLA, the first Vision-Language-Diffusion-Action model built upon pretrained diffusion-based Vision-Language Models (d-VLMs). In this paper, we explores how to effectively leverage d-VLMs for robotic manipulation. To address the domain gap between d-VLMs and the VLA, we propose a localized special-token classification strategy to reduce the adaptation difficulty and enable more efficient model training. Furthermore, to allow the masked diffusion paradigm to generate structured action trajectories, we introduce a hierarchical action-structured decoding strategy, which enables the model to produce action sequences in a hierarchical manner, resulting in more coherent and plausible outcomes. With these designs, LLaDA-VLA achieves state-of-the-art performance, demonstrating remarkable effectiveness across multiple simulation benchmarks as well as real-robot experiments. These results provides a solid foundation for exploring the application of d-VLMs in robotic manipulation and paving the way for future research in this direction.

{\small
\bibliographystyle{ieee_fullname}
\bibliography{egbib}
}
\end{document}